# Agents for Everyone: A Workshop Framework for Building Agentic AI Capabilities in a Distributed Curation Community

Seth Carbon[1], Sierra Moxon[1], Kimberly Van Auken[2], Pascale Gaudet[3], Christopher J. Mungall[1,*]

[1] Environmental Genomics and Systems Biology Division, Lawrence Berkeley National Laboratory, Berkeley, CA 94720, USA
[2] Division of Biology and Biological Engineering, California Institute of Technology, Pasadena, CA 91125, USA
[3] Swiss-Prot Group, SIB Swiss Institute of Bioinformatics, Centre Medical Universitaire, Geneva, Switzerland

* Corresponding author: cjmungall@lbl.gov

**ORCID:** S. Carbon: 0000-0001-8244-1536; S. Moxon: 0000-0002-8719-7760; K. Van Auken: 0000-0002-1706-4196; P. Gaudet: 0000-0003-1813-6857; C.J. Mungall: 0000-0002-6601-2165

## Abstract

Agentic AI has the potential to accelerate curation of biological databases and knowledge bases. However, uptake has been hindered by a number of challenges and obstacles, including access to agents and appropriate training. Here we describe how we have attempted to address and mitigate these challenges and obstacles through the deployment of a cloud-based agentic environment, and the development of an interactive training workshop for the Gene Ontology Consortium.

Our cloud environment for agentic-assisted curation was based on the JupyterHub platform, and utilized Claude Code as a universal harness. This allows curators to interact with an agent session through a terminal running in the browser, and has additional benefits such as centralization of access through a single API gateway, removing the need for participants to manage subscriptions or install software locally. We created four training modules, walking participants through basic agentic tool use first and then working up to agentic biological pathway curation using the existing GO-CAM (GO Causal Activity Model) curation tool.

Thirty-seven participants took part in the four-hour workshop. Our key takeaway from this workshop is that building community capability with agentic AI is primarily a problem of access, workflow design, and training. Removing technical barriers, introducing capabilities gradually, grounding exercises in familiar curation tasks, and giving curators direct experience evaluating agent output can provide a practical route toward building shared agentic AI capability in distributed scientific communities.

## Background: building AI capability in a distributed curation community

Research in biomedicine and the life sciences relies heavily on multiple data repositories, knowledge bases, and ontologies [1], all of which must be continuously curated to remain up-to-date and reliable. This curation is performed by dedicated professional biocurators who synthesize information from the literature and data analyses, and interpret them through the lens of their own expert domain knowledge [2]. One of the key knowledge bases in the biosciences is the Gene Ontology (GO) [3], which assigns functions to gene products across the tree of life, and curates causal activity models detailing the coordinated activities of gene products that execute biological programs [4]. The curators for the GO knowledge base are drawn from an international consortium, with representatives from different member knowledge bases, distributed across many different institutes. As in most biocuration projects, curation in GO is a collaborative exercise, involving coordination between member knowledgebases and the central coordinating body, known as GO Central.

The advent of artificial intelligence (AI) methods, and in particular agentic AI, presents an opportunity to assist curators and to accelerate the accumulation of knowledge [5]. Agent harnesses such as Claude Code were designed as tools for software development: they can inspect and edit files, execute commands, and interact with external tools. Their underlying file- and tool-oriented workflow is also applicable to broader forms of knowledge work; observed uses already include data analysis and producing prose-based documents, including work performed by people outside software occupations [6]. Although the potential application of agents is widespread, there is no single best way to integrate agents into existing workflows. Effective users adapt instructions, tools, permissions, and verification mechanisms to their particular tasks, and develop judgments about which tasks can be delegated and which require active oversight [7].

Despite the great potential of agentic AI to accelerate curation, as yet, there are a number of barriers and obstacles that impede uptake[8]. One of the main concerns is around reliability, with hallucinations (plausible-sounding but false AI-generated statements or references [9]) cited as a major worry. On top of this, there are many practical barriers to adoption, including difficulty installing software and difficulty navigating subscriptions to AI services through their institutions. Even if these practical barriers can be surmounted, there is a lack of agentic AI training documentation and materials aimed at biocurators.

Within the GO consortium, we decided to make a concerted effort to build up agentic AI capabilities across our network of biocurators. Our goal was not to introduce a particular curation tool or workflow, but to instead teach the fundamentals of agentic AI usage and to enable a community of practice around AI use. Accordingly, the specific agent harness that we used should be viewed as an implementation choice rather than the focus of the work. We sought to identify the environmental, teaching, and workflow-design choices that allow domain experts with heterogeneous technical backgrounds to gain practical experience with agents

We did this through two workshops. The first workshop, held in 2025, introduced the core concepts of generative AI and Large Language Models (LLMs). The materials for this workshop are online in Zenodo [10]. The second workshop, described here, was a hands-on guide to using agents.

## Development of the workshop environment and materials

### Use of Claude Code as agent harness

An agentic harness is a computer program that orchestrates an *agentic loop* and manages interactions with a user [11]. The user interacts with the harness through a chat-like interface, with the user providing instructions, and the harness performing the core loop, which iterates between *tool calls* and LLM calls. The ability of the system to call tools is a key feature that makes agentic AI more powerful than a chatbot. Tool calls are mediated through either the Model Context Protocol (MCP) [12] or through direct command line calls, often described using agent skill frameworks [13]. A growing number of MCPs and skills are available for agentic curation work, including the Ontology Lookup Service (OLS) MCP, and the AI4Curation skills collection [14].

For the workshop, we had the option of writing our own agentic harness (for example, using a framework such as LangGraph [15]), or using an off-the-shelf harness. We decided to use Claude Code (CC[1]) as the harness (specifically the command line/terminal version). CC is already fully featured, with well-tested integration of agent skills, compaction, and other features we knew we would need. We have also been using CC integrated with GitHub Actions as part of the GO ontology development workflow, via our dragon-ai-agent [16], and some GO ontology editors are already familiar with using CC directly.

Two characteristics of CC present challenges for use in a curation workshop. First, CC is primarily aimed at software developers, and employs user interaction modes and workflows that are optimized for this audience. We discuss mitigations for this first challenge later. The second, and more problematic, challenge is that CC is intended to be used locally on a user's desktop or laptop, along with any tools the agent needs to use. This presents challenges for local installation, as well as a lack of isolation between the agent's work environment and other files. Furthermore, the user must manage API keys or obtain a subscription. This presents major problems for deployment across multiple curators spread across multiple institutes.

[1] Although the coding agent is often shortened to simply "claude", this can also mean the web chat application, so we use the term "Claude Code" and the abbreviation CC to clearly denote that we are talking about the agentic harness.

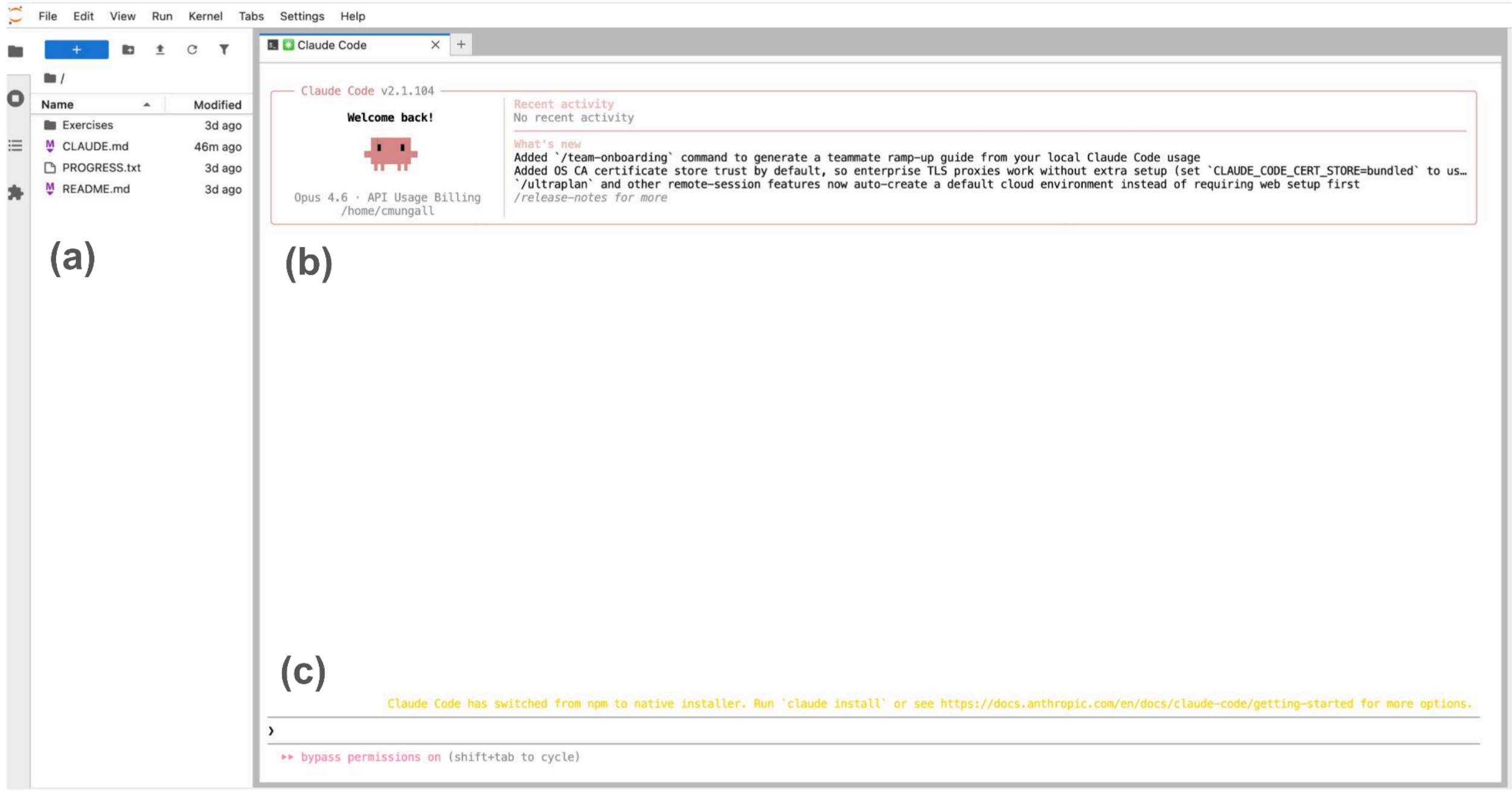

***Figure 1**: **The workshop environment as seen by participants in a web browser.** (a): JupyterLab file browser showing seeded files (Exercises, CLAUDE.md, PROGRESS.txt). (b): Claude Code running in a JupyterLab terminal (the Claude Code mascot icon is visible in the tab header); welcome screen is currently shown (c) text box for curator to enter instructions.*

### A cloud-based JupyterHub environment for agent use

Our approach was to host a shared JupyterHub [17] instance where every participant gets an isolated environment with CC pre-configured. This approach was inspired by the DOE BERIL (Biological and Environmental Research Integrated Laboratories) project [18], which uses KBase infrastructure [19] to provide shared computational environments for distributed scientific communities. In our installation, participants could all log into the shared instance with GitHub credentials they already had, after which they could use an instance of Claude Code running in the terminal in a browser window. Each participant had their own workspace, seeded with the same tutorial materials. **Figure 1** shows the participant view. See **Appendix A** for a full set of design criteria for the environment.

The difference between the cloud-based environment we provided and a standard local setup can be seen in **Figure 2**.

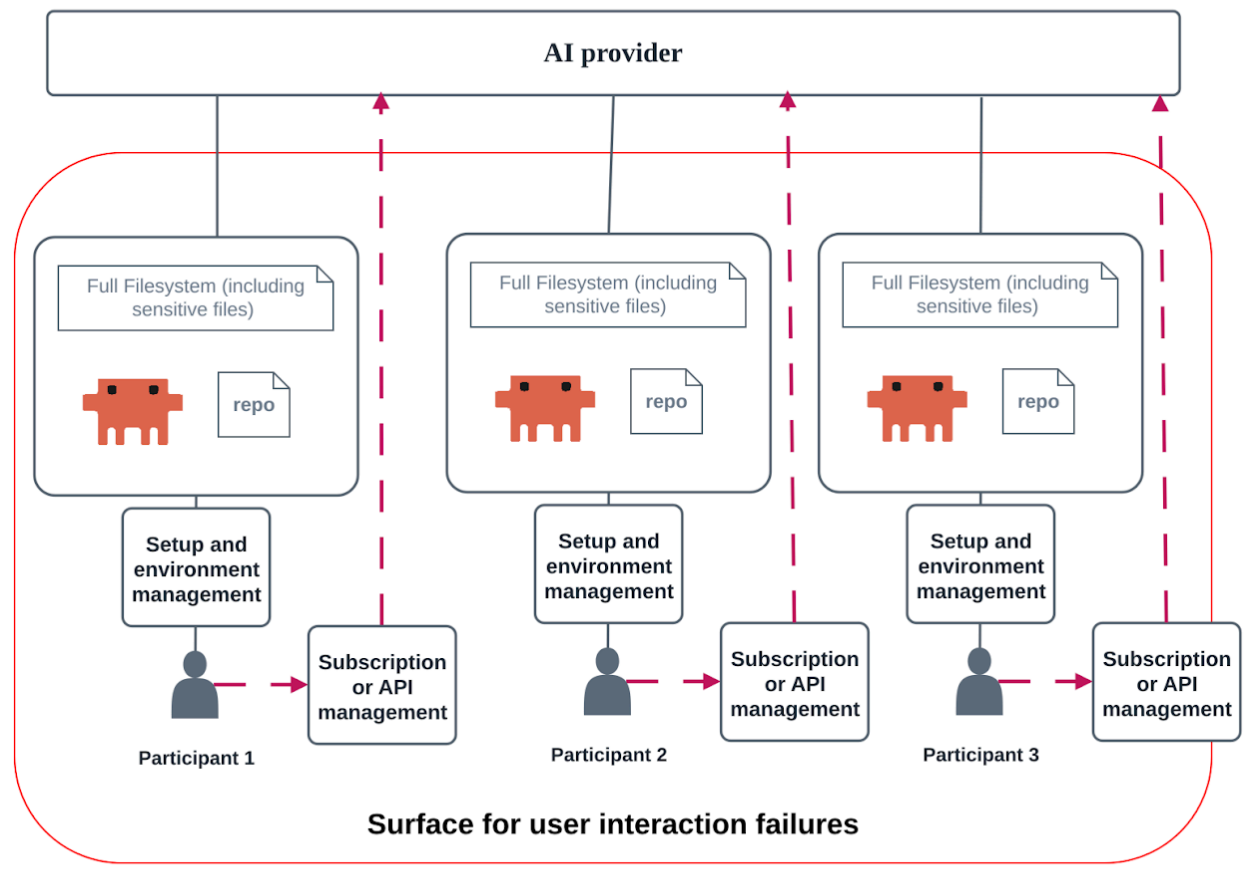

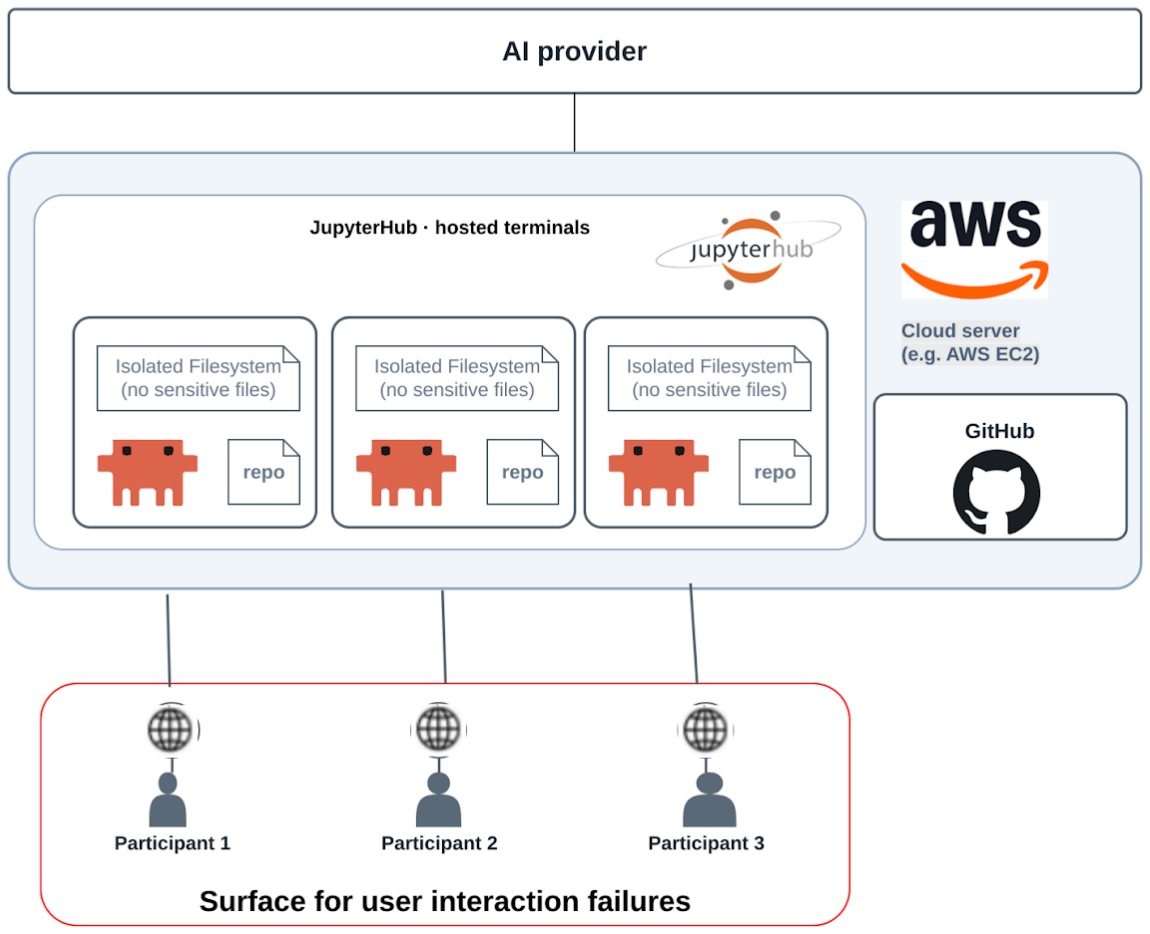

***Figure 2**: **Architecture comparison**. (a) Individual installation on local machines. Agent harnesses such as CC are installed locally: each participant needs to manage their subscription or API keys, local environment, repo checkout, and installation. Agents have the ability to read/write files on the entire system with the same permissions as the controlling user. (b) (Our approach) Shared JupyterHub model: a single cloud instance with hosted terminals, a single shared API credential, and GitHub-based authentication. Agent harness is pre-installed and executed on startup. Each user has their own local file space and repo checkout, isolated from other files on their local machine. With most resources being managed remotely, thereby reducing the surface for interaction failures, the workshop participant only needs access to their web browser to fully participate and explore.*

## The tutorial structure

We designed this workshop as a series of four exercises, each building on the previous one and introducing new agent capabilities. The workshop started with a short session introducing the conceptual foundations of agentic AI, but most of the session was interactive, with an instructor walking participants through each exercise: first demonstrating their use, then having each curator perform the same exercise for themselves in their JupyterLab cloud session. We also created an agent skill that was instructed to provide individualized help to each participant.

**Table 1**: Workshop exercises.

| Exercise | What participants do | Aim and tools |
|---|---|---|
| **1. Getting started** | Chat with the agent; create, edit, and translate files | Comfort with terminal and conversational interface<br>*Tools: file I/O only* |

| Exercise | What participants do | Aim and tools |
|---|---|---|
| **2. Literature and ontologies** | Search PubMed for papers on Epe1 (S. pombe); annotate GO terms in abstracts; query existing GO annotations; fetch PDFs | External APIs and MCP (Model Context Protocol) servers; structured output (CSVs); critical evaluation of retrieved data *Tools: NCBI (National Center for Biotechnology Information) E-utilities, OLS, QuickGO* |
| **3. Scripting with PANTHER** | Look up gene families and orthologs; direct the agent to write a Python script comparing MF (molecular function) annotations across species | Agent-as-programmer: participants direct script creation without writing code themselves. Importantly, the gated JupyterLab cloud environment keeps code written by agents separate from the user's desktop environment. *Tools: PANTHER API* |
| **4. GO-CAM model review** | Review an incomplete GO-CAM model; identify missing molecular functions; find evidence; build a corrected model | Capstone integrating all prior skills; critical evaluation against domain expertise *Tools: Noctua/Barista MCP, PubMed, OLS* |

The set of exercises is shown in **Table 1**, and fully detailed in **Appendix B**. The first exercise walked curators through basic interactions with an agent, where the agent makes simple changes to files within the participants' workspace. While not markedly different from a typical online chat session, this was intended to introduce simple tool use (here, Unix file editing and file operations) in an agentic session.

The second exercise focused on using external APIs through tools, particularly Model Context Protocol (MCP) services, a method for providing tools. We used the Ontology Lookup Service (OLS) MCP service [20] to search for and retrieve ontology terms, and the PubMed [21] MCP service to find papers and retrieve abstracts. In this context, tools wrapping resources such as OLS and PubMed provide opportunities to ground and verify model outputs, including reducing errors in ontology term identifiers [22].

The third exercise aimed to address a common curatorial need to employ basic programming and/or SQL to perform ad-hoc queries for data formatting or reporting. In this exercise, participants learned how to direct an agent to write Python scripts against the PantherDB [23] API endpoint to retrieve sets of annotations for homologous genes.

In the fourth and final exercise, we developed an agent skill in advance that communicated with the GO-CAM web-based curation tool 'Noctua' [24], wrapping the existing API the current user interface (UI) uses. This enabled the agent to both read and write over actual curated pathways. To avoid potential changes to production data, we wired it to communicate only with the non-production server. During the exercise, curators were encouraged to search for pathways and to update example pathways interactively via natural language. In addition to the standard CC chat interface, they could see pathway diagrams being built by the agent in the standard, familiar Noctua UI view and make edits alongside agentic edits in this view. We call this the “shared control” or “sidecar” pattern (Figure 3).

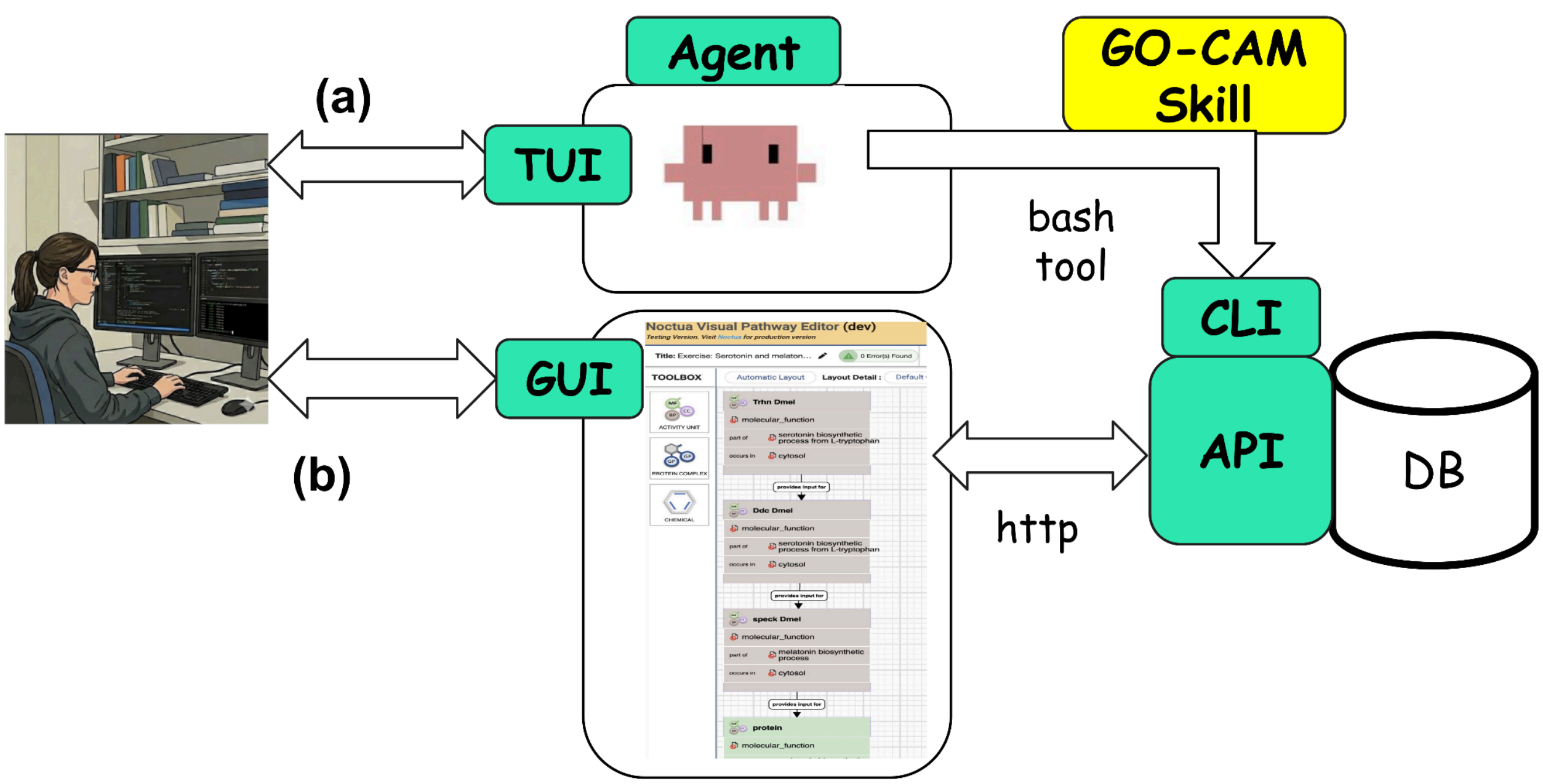

***Figure 3**. **Shared control (“sidecar”) pattern for accessing curation interface.** (a) Curator interacts with agent via text-oriented Text User Interface (TUI), agent uses skill to learn command line interface (CLI) which wraps Application Programmer Interface (API). The API reads from and updates the shared database (DB). (b) Curator interacts with DB via a Graphic User Interface (GUI). Updates from (a) are reflected directly in the GUI, so the curator is able to see the results of agent actions in real time; the curator can simultaneously edit in GUI while directing agent via agentic TUI.*

## Workshop outcomes

The workshop took place on April 14, 2026, with 37 active participants across a 4-hour session. Over half the participants completed all four exercises during the workshop. An anonymous survey conducted after the workshop revealed that most participants felt the duration and level of technical detail were appropriate, and that the material was directly relevant to their work. Here we present some general highlights and discussion points. A detailed set of statistics is available in **Appendix D**.

### JupyterHub mitigates many obstacles to local agentic AI use

Because we ran the workshop exercises inside JupyterHub, participants did not have to spend any time on installation or local setup. They opened a URL, logged in with GitHub, and were placed directly in a CC terminal session inside their web browser. Several participants commented that they would not have been able to install CC locally due to IT restrictions at their institution. Additionally, participants did not have to get Claude accounts, obtain subscriptions, or obtain API keys. All participants were automatically configured to use a central API key administered by the workshop organizers.

We considered other options for the workshop. For example, Claude Code on the Web (claude.ai/code) provides an intuitive web-based agentic interface in a cloud environment, but would have required every participant to have their own subscription. For a previous workshop [25] organized as a part of the International Conference on Biomedical Ontologies (ICBO), we used GitHub Codespaces [26]. This provides a similar experience to JupyterHub, with a terminal in a browser, but has a number of downsides, including longer startup times (up to 5 minutes) and, crucially, no secure way to centralize token-based access to the aent.

While overall we were happy with the selection of the JupyterHub framework, we discovered some usability issues due to how it interacts with CC. Some of these can be mitigated, but some remain unsolved. For example, during testing, we identified an annoying issue with terminal flickering, which we mitigated during the workshop. Other issues, such as difficulty copying text from the terminal, are things we'd still like to address in future iterations.

Another aspect we were unsure about before the workshop was the use of an agentic harness aimed at developers. During beta testing, we identified a number of "cute" features in coding agents, such as random "spinner tips" (a collection of verbs that are shown in the terminal during LLM calls), that we found distracting and confusing, so we created a minimal configuration that switched these off. One thing we did not catch during beta testing was that coding agents naturally try many different command-line options iteratively, which frequently results in tool failures that appear as errors in the terminal. This is, in fact, how agents are meant to work, but many participants naturally interpreted this as more serious errors, as participants may see the error, but not necessarily the recovery. A full list of potential barriers and mitigations is listed in **Appendix C**.

However, overall, we felt these downsides were outweighed by the benefits of a completely open-ended, configured agent environment in an isolated space, without any worries about installation or subscribing to AI services.

### Using agentic interfaces alongside existing curation tools adds relevance

We intentionally built up the exercises slowly, gradually introducing core concepts like agentic tool use. The first three exercises were aimed to be more pedagogical than practical, with the fourth and final exercise introducing an actual curation workflow. This last exercise involved using an existing curation interface, with the agent given read/write access via an MCP (Figure 3). In contrast to a purely terminal-based curation experience, curators could see their model being modified by the agent in a separate browser window.

Participants found this exercise the most relevant to their work, and they were able to put into practice lessons learned from previous sections. 35 of 37 participants (95%) made at least one agentic call to Noctua, with over a thousand Noctua-related tool invocations. Nine participants created entirely new GO-CAM models during the workshop, going beyond the planned exercises.

**Agentic workflows should be coupled with strong guardrails**

One of the major barriers to AI adoption in biocuration is concerns about reliability, particularly hallucinations. Many people using AI for data and curation tasks have experienced guiding a ChatGPT session in which plausible-sounding references or ontology term IDs are returned that turn out to have been fabricated by the LLM.

We have found that combining two different strategies is optimal for mitigating hallucination. The first strategy is **prospective**: providing agents with tools to perform lookups (e.g. lookups of ontology terms from OLS or papers from PubMed), in the form of MCPs or skills. Studies have shown that when agents are given the opportunity to look up something rather than rely on latent memory, they are less liable to hallucinate ontology term identifiers or citations [22]. The second strategy is **retrospective**: use deterministic procedures to check the output of the agent at each turn, but before exiting the agentic loop. We made heavy use of the LinkML [27] suite of tools, in particular the LinkML term validator [28] and the LinkML reference validator [29], as powerful guardrails to mitigate hallucination.

We find it most effective to combine both prospective and retrospective strategies, but for simplicity we did not build the second mechanism into the workshop, thinking the MCP approach would be sufficient.

During the second workshop exercise, in a step where the agent was requested to add abstracts from a list of annotations (including PMIDs) to a table, the expectation was that the agent would use the PubMed MCP to look up all publications using the PMIDs. However, many participants noticed that the abstracts added were plausible but not the real abstracts. After the workshop, we examined the agent traces and observed that the PubMed MCP server occasionally failed to authenticate on initialization (perhaps due to multiple simultaneous users). In these cases, the agent resorted to a separate API, which truncated the results. The agent then “helpfully” filled in the missing values in these truncated results using its training data. Importantly, it did this without explicitly informing the user — a failure pattern known as weak fallback or silent degradation, where the agent enters a degraded operating mode without surfacing the downgrade [30].

The lesson is to take extra care to mitigate both hallucinations and weak fallback in the agent harness configuration.

# Conclusions

Agentic AI is a powerful tool, but effective use depends on domain experts driving it. Enabling domain experts to use agents effectively can be challenging because of obstacles to accessing agents and a lack of appropriate training. We devised a training environment and tutorial materials based on the JupyterHub platform to mitigate many of these challenges. Using this framework, we ran a half-day workshop for Gene

Ontology Consortium curators. Preliminary feedback suggests the workshop successfully provided access and conceptual grounding in agentic AI. The workshop also demonstrated the possibility of using agentic AI alongside existing curation tools in a “sidecar” pattern, where the agent operates alongside the curator’s existing interface rather than replacing it. We plan to continue to use our platform for continued training and curation.

# Author contributions

SC designed, implemented, and deployed the workshop infrastructure and framework (Terraform, JupyterHub, authentication). SM and SC developed the budget-enforcer software and configured AI model access via Vertex AI. PG, KVA, and CJM developed the workshop content and exercises. CJM conceived and led the overall workshop design. All authors contributed to the manuscript.

## Acknowledgements

The GO Consortium is funded by the National Human Genome Research Institute (US National Institutes of Health), grant number HG012212, with co-funding by NIGMS. CM, SC, and SM are also supported by the Director, Office of Science, Office of Biological and Environmental Research, of the U.S. Department of Energy under Contract No. DE-AC02-05CH11231. We thank Gazi Mahmud (KBase, Lawrence Berkeley National Laboratory) for advice on setting up JupyterLab for multi-user workshops, Andrew Schmeder for assistance with Vertex AI configuration, and Jonah Cool and the Anthropic AI4Science program for ongoing support of agentic access for curators. We are grateful to all workshop participants for their engagement and feedback.

# Appendices

## Appendix A: Design Decisions for Workshop Environment

| Design Decision | Description |
|---|---|
| Isolated/sandboxed user environments | Each participant gets their own Unix account and home directory on the server. Users cannot see each other’s work, but the infrastructure is simple — no containers or Kubernetes. In contrast, an agent harness running locally on a curator’s home or work machine would have access to all files (including the ability to delete). |
| GitHub authentication | GO Consortium members already have GitHub accounts (already required for some consortium activities), so GitHub OAuth was a natural fit. The consortium’s existing `users.yaml` authorization registry (originally created for Noctua access control) served as the allowlist, requiring no new user management infrastructure. |
| Shared API credentials | A single Google Cloud Platform (GCP) service account provides access to Anthropic models via Vertex AI, with budget enforcement on the GCP side. Participants never see or manage API keys. |
| Standardized agent configuration | Each user’s home directory is populated from a template containing a `CLAUDE.md` (instructions that shape the agent’s behavior), exercise files, domain-specific skills, and configuration files that suppress onboarding prompts and other UI noise (see barrier taxonomy below). |
| Minimal obstructions to starting work | When users open a terminal, they see a simple menu: they press Enter to launch Claude, or T for a plain shell. |

## Appendix B: Exercise descriptions

### Exercise 1: Getting started

Participants are introduced to the terminal-based chat interface without any domain-specific content. They chat with the agent, ask it to create files (e.g. fun facts about a city), edit and translate those files, and observe the results in the JupyterLab file browser. The goal is basic comfort: understanding that the agent can read and write files, that commands are conversational, and that the file browser shows results in real time. No biological knowledge is required.

### Exercise 2: Literature and ontology tools

Participants search PubMed for papers about the histone demethylase Epe1 in *Schizosaccharomyces pombe*, a well-studied fission yeast gene chosen because it has a manageable literature (tens of papers, not thousands). The agent uses NCBI E-utilities to retrieve papers, then the Ontology Lookup Service MCP server to identify GO terms mentioned in abstracts. Participants create CSVs of results, query existing GO annotations for the gene, and attempt to fetch full-text PDFs. The exercise deliberately surfaces the abstract fabrication issue (see Results) as a teaching moment about verifying agent output.

### Exercise 3: Scripting with PANTHER

Participants ask the agent to look up Epe1 in the PANTHER protein classification database, find human orthologs, and write a Python script that compares molecular function annotations across species. The key pedagogic shift: participants direct the creation of code without writing it themselves. They evaluate the output, request improvements (add gene symbols, include ortholog types, try different genes from their own organism of interest), and critically compare PANTHER's automated annotations against curated GO annotations.

### Exercise 4: GO-CAM model review and construction

The capstone exercise. Participants are given an incomplete GO-CAM model of the Drosophila serotonin and melatonin biosynthesis pathway, pre-loaded on the Noctua development server. The model contains four activities with gene products, biological processes, and a causal chain — but all molecular functions are set to the root term GO:0003674 ("molecular_function"), meaning "unknown." Participants must identify the correct molecular functions for each activity, find supporting evidence (PMIDs), and either plan or execute the model edits via the Noctua/Barista API. This exercise requires integrating all skills from previous exercises: literature search, ontology lookup, and API interaction, guided by domain expertise.

## Appendix C: Configuration of JupyterHub

### Lessons from dry runs

We ran several test sessions before the actual workshop. Every one surfaced issues that would have been disruptive in a real session. We categorize these as **barriers** — obstacles that prevent or degrade the workshop experience — and document both the barrier and our mitigation. Technical details of each mitigation are below.

| Category | Barrier | Impact | Mitigation |
|---|---|---|---|
| **Terminal UX** | Scrollbar resets during streaming | Participants can't scroll back | `CLAUDE_CODE_NO_FLICKER=1` (partial fix) |
| **Terminal UX** | Can't copy text from terminal | Frustrating for power users | Tradeoff: mouse capture needed for scrolling (UNSOLVED) |
| **UI noise** | Tips, surveys, onboarding prompts | Confuses non-technical users | Pre-seed settings to suppress all prompts |
| **UI noise** | Auto-update error messages | Alarming on startup | `DISABLE_AUTOUPDATER=1` |
| **UI noise** | Workspace trust dialog | Blocks first launch | Pre-seed `~/.claude.json` with trust accepted |
| **Auth** | Open registration insecure | Anyone with URL can join | GitHub OAuth gated by consortium user registry |
| **Tool Use** | Permission requested for all tool use | Non-technical users cannot assess impact of arbitrary commands | Run in isolated environments in skip permissions mode |
| **Tool Use** | Red error text during iteration | Curators think something is broken | Explain in exercises; CLAUDE.md guidance |
| **Harness** | Premature compaction | Performance degradation; agent "forgets" | Instructions to `/clear` at end of each exercise |
| **LLM** | Poor LLM responses in agent loop | Agent is "less smart" | Default to highest quality model (opus) |
| **LLM** | Hallucinations | Cryptic errors; loss of trust | Prospective: avoid LLM-as-Oracle; tool use for look ups<br>Guardrails: LinkML validator hooks |
| **Scaling** | ~500MB RAM per user session | System thrashing at 10+ users on 8GB | Memory-optimized cloud instances |
| **Scaling** | External API overload (PANTHER) | 40 concurrent users = accidental DDoS | Stagger exercises; use resilient APIs |
| **Agent tone** | Informal language defaults | Mismatch with professional context | CLAUDE.md tone instructions |

**Terminal scrollbar resets**

Claude Code's TUI redraws the terminal aggressively — erasing lines and repositioning the cursor on every output chunk. In a native terminal this is invisible, but JupyterLab's terminal (built on xterm.js) responds by resetting the scrollbar to the top.

Setting CLAUDE_CODE_NO_FLICKER=1 (an environment variable added in Claude Code v2.1.89) uses an alternate rendering path with fewer cursor-repositioning sequences. This reduces but doesn't fully eliminate the problem. See anthropics/claude-code#36128, #34845, #34400.

A related tradeoff: CLAUDE_CODE_DISABLE_MOUSE=1 disables mouse capture, allowing text selection/copy in the terminal — but breaks scrolling. We chose to keep mouse capture (scrolling) and accept that copy-paste requires writing to files.

### Suppressing UI noise

Claude Code shows rotating spinner tips, feature announcements, feedback surveys, and onboarding hints. We suppressed these via:

**Documented settings** in `~/.claude/settings.json` (seeded by the `pre_spawn_hook`): - `spinnerTipsEnabled: false` - `feedbackSurveyRate: 0` - `skipDangerousModePermissionPrompt: true`

**Undocumented internal state** in `~/.claude.json` (seeded per-user, since the project path must match the user's home directory): - `hasCompletedOnboarding: true` - `hasTrustDialogAccepted: true` (per-project) - `hasCompletedProjectOnboarding: true` - High counters on `opus1mMergeNoticeSeenCount`, `ideHintShownCount`, `voiceNoticeSeenCount`, `numStartups` - `tipsHistory` pre-populated

**Environment variable**: `DISABLE_AUTOUPDATER=1` prevents the auto-update check that fails (and alarms users) when Claude Code is installed system-wide and individual users can't write to `/usr/lib/node_modules/`.

All seeding is done atomically (write to temp file, fsync, rename) in the `pre_spawn_hook` to prevent truncated JSON if the spawner is interrupted.

### Authentication configuration

We iterated through three au¥thentication approaches:

1. **`FirstUseAuthenticator`** — users pick username/password on first login. Zero admin overhead but insecure (anyone with the URL can register). Used during early development.
2. **Allowlist file** — `allowed_users.txt` restricts who can register. Simple but requires manual curation.
3. **Bimodal auth via `jupyterhub-multiauthenticator`** — GitHub OAuth (Tier A, gated by `go-site/metadata/users.yaml`) plus PAM (Tier B, for workshop-specific accounts). The login page shows a chooser. Cloud-init fetches `users.yaml` at boot and runs `scripts/extract_github_users.py` to populate the OAuth allowlist.

The multiauthenticator implementation surfaced bugs around form-based sub-authenticators (PAM) coexisting with OAuth-based ones.

**Memory and scaling**

Each active Claude Code session uses ~450-500MB RAM (`Node.js` process + JupyterHub single-user server). Planning numbers:

| Users | Minimum RAM | Recommended |
|---|---|---|
| 10 | 8 GB | 16 GB |
| 25 | 16 GB | 32 GB |
| 50 | 32 GB | 64 GB |

CPU load peaked above 3.0 during Exercise 3 when participants ran Python scripts and external API calls concurrently. While Claude Code is mostly I/O bound (waiting on API responses), script execution and `uv` tool installations are CPU-bound and can spike simultaneously. A compute-optimized instance family (e.g. `c6i`) with more vCPUs and less RAM may be a better fit than the memory-optimized `r6i` we used. Disk should be 150-200GB to accommodate per-user tool caches.

**Agent safety mitigations**

- Noctua API tokens restricted to the dev/test server (not production)
- GCP budget enforcement via `budget-enforcer` caps Vertex AI spend. This is essential for running workshops without open-ended billing risk.

**Full setup**

The full setup is open source at https://github.com/geneontology/go-jupyter. Key files:

| **File** | **Purpose** |
|---|---|
| `jupyterhub_config.py` | Hub config: authentication, spawner, Vertex AI env vars, user seeding |
| `terraform/` | AWS infrastructure (EC2, DNS, TLS, cloud-init) |
| `scripts/extract_github_users.py` | Extracts GitHub usernames from users.yaml for the OAuth allowlist |
| `user-template/` | Files seeded into each user's home |
| `user-template/CLAUDE.md` | Tutor persona and exercise instructions |
| `user-template/.bashrc` | Welcome screen with Claude/shell choice |
| `systemd/jupyterhub.service` | systemd unit for the hub |

**Final setup**

- **Instance**: AWS `r6i.8xlarge` (256GB RAM, 32 vCPUs) in `us-east-1`
- **Model**: Claude Opus 4.6 via Vertex AI (we opted for the most capable model to give participants the best experience)
- **Auth**: GitHub OAuth + PAM side door via `jupyterhub-multiauthenticator`
- **Cost**: ~$388 total — $8 for the EC2 instance ($2/hr × 4hrs) plus $379.70 in Vertex AI API charges ($374.20 for Opus 4.6, $5.50 for Haiku 4.5 subagent calls; Figure 5)

- **URL**: https://jupyter-test-<hash>.geneontology.io (auto-generated hostname with Let's Encrypt TLS)

## Appendix D: Workshop statistics

### Participant progress

Each user's agent maintained a PROGRESS.txt file tracking exercise completion. We analyzed these after the workshop:

| Furthest exercise reached | Participants | % of active |
|---|---|---|
| All 4 exercises | 21 | 57% |
| Exercise 3 (PANTHER scripting) | 10 | 27% |
| Exercise 2 (PubMed + OLS) | 4 | 11% |
| Exercise 1 (basics) | 2 | 5% |

**Over half the participants completed all four exercises in a 4-hour session.** The main drop-off point was between Exercise 3 and 4.

### Tool call analysis

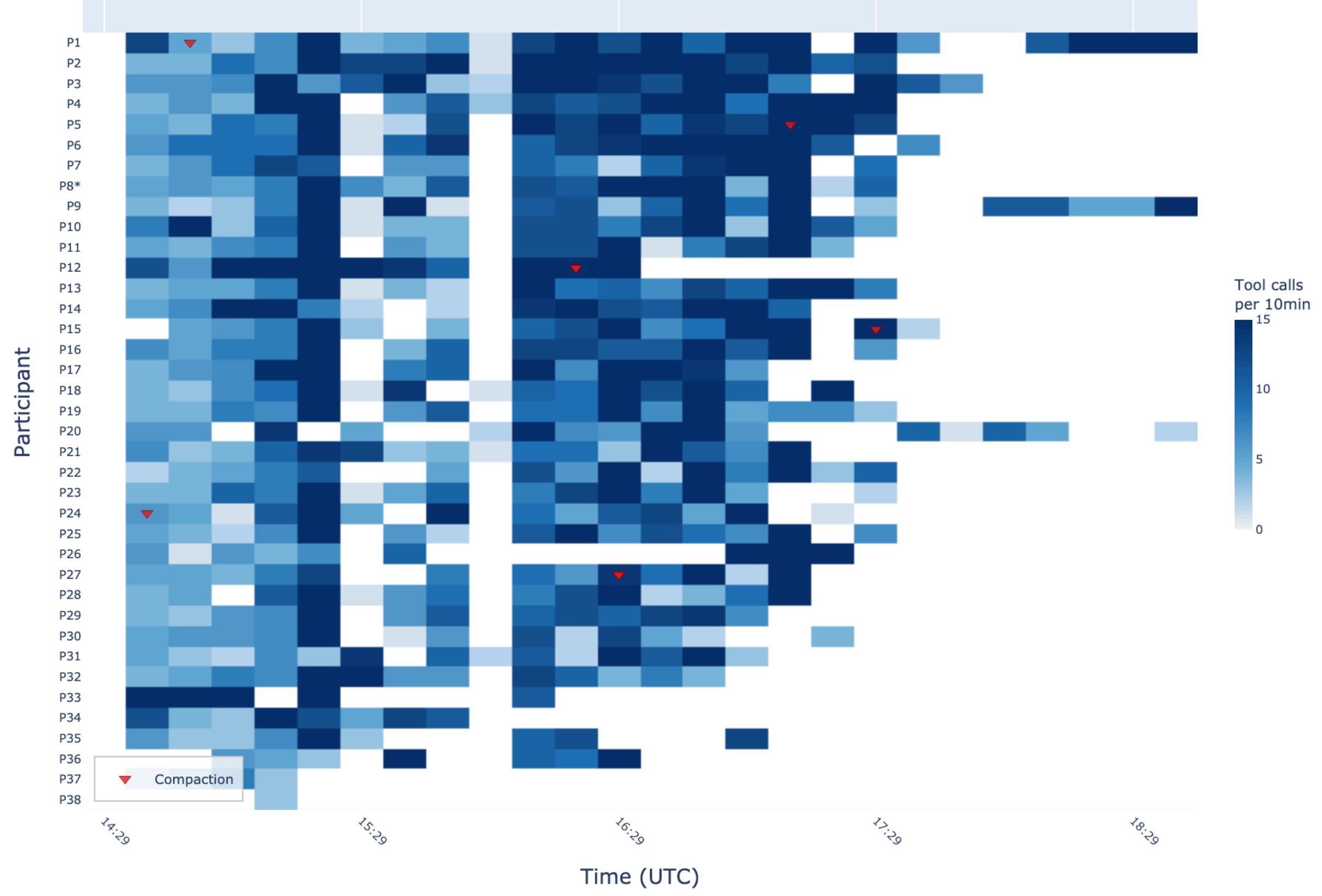

***Workshop activity timeline.*** Each row is an anonymized participant (P8* = instructor), sorted by total engagement. Color intensity shows tool calls per 10-minute bin. Red triangles mark context compaction events. The white band around 15:30-16:00 UTC corresponds to the live demonstration period. Late joiners (P9, P33-P38) and early leavers are visible at the edges.

## Resource usage

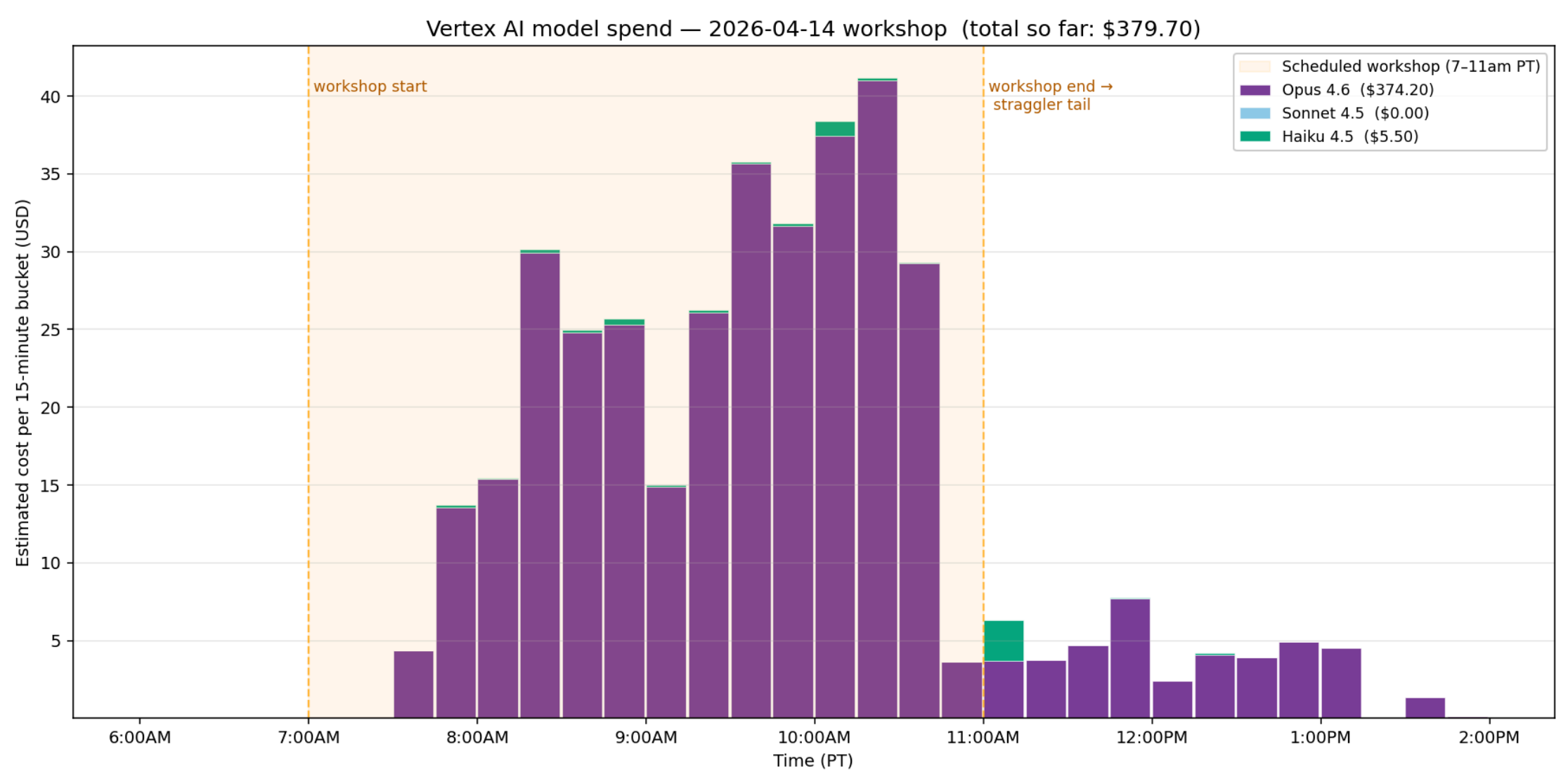

*Vertex AI model spend during the workshop, broken down by model and 15-minute time buckets.* Total API cost was $379.70, dominated by Opus 4.6 ($374.20) with a small contribution from Haiku 4.5 subagent calls ($5.50). Peak spend reached ~$40 per 15-minute bucket around 10:00 AM PT. The straggler tail after the scheduled end shows participants continuing to work independently.

## Agent behavior analysis

We analyzed the agent's tool usage and failure modes across all workshop sessions by parsing Claude Code session logs and classifying agent actions using a taxonomy of agentic failure modes [30]. This analysis focuses on the *agent's* behavior — what tools it invoked, how often it erred, and what failure patterns emerged — rather than on participant interactions.

**Aggregate numbers** across 153 session trace files:

| Metric | Value |
|---|---|
| Total agent tool invocations | 5,816 |
| Average tool calls per session | ~38 |
| Sessions with classified failure modes | 75 (49%) |

**Tool usage breakdown:**

| Tool | % of calls | Notes |
|---|---|---|
| Bash | 47% | curl (PubMed, PANTHER), noctua CLI, Python scripts |
| OLS MCP | 13% | Ontology term lookups — curators' natural instinct |
| Read | 13% | Reading files, exercise instructions |
| Write/Edit | 14% | Creating CSVs, scripts, model files |
| WebFetch | 4% | Fetching PDFs and web content |
| Skill | 3% | /exercise commands |

The OLS MCP being the second most-used tool (after Bash) was a pleasant surprise — curators naturally reached for ontology term lookup, which is exactly the behavior we'd want to see.

**Failure mode classification:**

| Failure mode | Count | Interpretation |
|---|---|---|
| tool_misuse | 57 | Agent tried a tool with wrong arguments or got an error |
| premature_termination | 30 | Agent stopped before validating completion |
| context_limit | 13 | Hit context window limits |
| state_desynchronization | 1 | Context loss after compaction |

The **tool error rate was 1%** (57 misuse events out of 5,816 tool calls) — remarkably low for non-technical users exploring unfamiliar APIs. Most tool errors occurred when the agent tried an API call, got an error response, and self-corrected on the next attempt. This is normal agentic behavior, but it maps to the "red error text" problem described earlier — curators see the error, not the recovery.

**Context window limits and compaction.** Claude Code manages a finite context window (1M tokens for Opus 4.6). When the context fills, it triggers *compaction* — older tool outputs are cleared, and the conversation is summarized to free space. We found 71 compaction events across 32 of 37 active participants (86%), indicating that most participants hit context pressure at some point during the 4-hour session. Three participants exhausted their context entirely, triggering a session continuation with the message "This session is being continued from a previous conversation that ran out of context." This is significant because compaction can cause loss of earlier instructions, exercise context, and intermediate results. The high compaction rate suggests the exercises are right-sized for a single session, but participants who go deep into any one exercise—or who do not start fresh sessions between exercises—should expect context-related degradation. We provided a /compact command for participants to use between exercises to proactively free context, but compliance was uneven — many participants simply continued without compacting, accumulating context across all four exercises.

**Premature termination** (30 events) was mostly a workshop artifact — participants moving between exercises or closing terminals, not a real agent failure.

**No security findings.** No unauthorized production writes, no prompt injection attempts, no data exfiltration. The `CLAUDE.md` instruction to use the dev server only was respected by the agent across all sessions.